\documentclass[10pt,twocolumn,letterpaper]{article}

\usepackage[utf8]{inputenc}
\usepackage{iccv}
\usepackage{times}
\usepackage{epsfig}
\usepackage{graphicx}
\usepackage{amsmath}
\usepackage{amssymb}

\usepackage[breaklinks=true,bookmarks=false]{hyperref}

\iccvfinalcopy 

\def\iccvPaperID{****} 

\ificcvfinal\fi

\begin{document}

\title{Understanding and Comparing Deep Neural Networks\\ for Age and Gender Classification}

\author{Sebastian Lapuschkin\\
Fraunhofer Heinrich Hertz Institute\\
10587 Berlin, Germany\\
{\tt\small sebastian.lapuschkin@hhi.fraunhofer.de}
\and
Alexander Binder\\
Singapore University of Technology and Design\\
Singapore 487372, Singapore\\
{\tt\small alexander\_binder@sutd.edu.sg}
\and
Klaus-Robert Müller\\
Berlin Institute of Technology\\
10623 Berlin, Germany\\
{\tt\small klaus-robert.mueller@tu-berlin.de}
\and
Wojciech Samek\\
Fraunhofer Heinrich Hertz Institute\\
10587 Berlin, Germany\\
{\tt\small wojciech.samek@hhi.fraunhofer.de}
}
\maketitle
\thispagestyle{empty}

\begin{abstract}
Recently, deep neural networks have demonstrated excellent performances in recognizing the age and gender on human face images.
However, these models were applied in a black-box manner with no information provided about which facial features are actually used for prediction 
and how these features depend on image preprocessing, model initialization and architecture choice.
We present a study investigating these different effects.

In detail, our work compares four popular neural network architectures, studies the effect of pretraining, evaluates the robustness of the considered alignment preprocessings via cross-method test set swapping and intuitively visualizes the model's prediction strategies in given preprocessing conditions using the recent Layer-wise Relevance Propagation (LRP) algorithm. 
Our evaluations on the challenging Adience benchmark show that suitable parameter initialization leads to a holistic perception of the input, compensating artefactual data representations. With a combination of simple preprocessing steps, we reach state of the art performance in gender recognition.

\end{abstract}

\section{Introduction}
\label{sec:introduction}

Since SuperVision \cite{krizhevsky2012imagenet} entered the ImageNet \cite{russakovsky2015imagenet} challenge in 2012 and won by a large margin, much progress has been made in the field of computer vision with the help of Deep Neural Networks (DNN). Improvements in network architecture and model performance have been steady and fast-paced since then \cite{zeiler2014visualizing,simonyan2014very,szegedy2015going,szegedy2016inception}. The use of artificial neural networks also has revolutionized learning-based approaches in other research directions beyond classical computer vision tasks, \eg by learning to read subway plans \cite{graves2016hybrid}, understanding quantum many-body systems \cite{schutt2017quantum}, decoding human movement from EEG signals \cite{StuJNM16,schirrmeister2017deep} and matching or even exceeding human performance in playing games such as Go \cite{silver2016mastering}, Texas hold'em poker \cite{moravvcik2017deepstack}, various Atari 2600 games \cite{mnih2015human} or Super Smash Bros.~\cite{firoiu2017beating}.

Automated facial recognition and estimation of gender and age using machine learning models has held a high level of attention for more than two decades \cite{kwon1994age,o1997sex,baluja2007boosting,guo2009study,gao2009face} and has become ever more relevant due to the abundance of face images on the web, and especially on social media platforms.
The introduction of DNN models to this domain has largely replaced the need for hand crafted facial descriptors and data preprocessing considerably increased possible prediction performances at an incredible rate. DNN models have been not only successfully applied for age and gender recognition, but also for the classification of emotional states \cite{ArbGCPR16}.
In the previous three years alone, age recognition rates increased from $45.1\%$ \cite{eidinger2014age} to $64\%$ \cite{rothe2016deep} and gender recognition rates from $77.8\%$ to reportedly $91\%$ \cite{dehghan2017dager} on the recent and challenging Adience benchmark \cite{eidinger2014age}, mirroring the overall progress on other available benchmarks such as the Images of Groups data set \cite{gallagher_cvpr_09_groups}, the LFW data set \cite{LFWTech} or the Ghallagher Collection Person data set \cite{gallagher_cvpr_08_clothing}.

Next to the indisputable performance gains across the board, the probably most important factor for the popularity of DNN architectures is the low entry barrier provided by intuitive and generic (layer) building blocks, the one-fits-all applicability to many learning problems and most importantly the availability of highly performing and accessible software for training, testing and deployment,  \eg Caffe \cite{jia2014caffe}, Theano \cite{2016arXiv160502688short}, and Tensorflow \cite{tensorflow2015-whitepaper}, to name a few,  supported by powerful GPU-Hardware.

However, until recently, DNNs and other complex, non-linear learning machines have been used in a black-box manner, providing little information about which aspect of an input causes the actual prediction. Efforts to \textit{explaining} such complex models in the near past have resulted in several approaches and methods \cite{zeiler2014visualizing,zintgraf2016new,ribeiro2016should,gevrey2003review,baehrens2010explain, simonyan2013deep,bach2015pixel} allowing for insights beyond the performance ratings obtainable on common benchmarks. This is a welcome development, as in critical applications such as autonomous driving or in the medical domain, it is often of special importance to know why a model decides the way it does, given a certain input, and whether it can be trusted outside laboratory settings \cite{lapuschkin2016analyzing}.

In this paper, we compare the influence of model initialization with weights pretrained on two real world data sets to random initialization and analyze the impact of (artefactual) image preprocessing steps to model performance on the Adience benchmark dataset for different recent DNN architectures.
We can show that suitable pretraining can yield a robust set of starting model weights, compensating artefactual representation of the data, via cross-method test set swapping.
Using Layer-wise Relevance Propagation \cite{bach2015pixel}, we visualize how those choices made prior to training affect how the classifier interacts with the input on pixel level, \ie how the provided input is used to make a decision, and what parts of it. We rectified the performance of \cite{rothe2016deep} on gender recognition referred to in \cite{dehghan2017dager} with a more likely result and report our own result, slightly exceeding that baseline. Via a combination of simple preprocessing steps, we can reach state of the art performance on gender recognition from human face images on the Adience benchmark dataset.

\section{Related Work}
\label{sec:relatedwork}

One of the more recent face image data sets is the Adience benchmark \cite{eidinger2014age}, which has been published in 2014, containing 26,580 photos across 2,284 subjects with a binary gender label and one label from eight different age groups\footnote{(0-2, 4-6, 8-13, 15-20, 25-32, 38-43, 48-53, 60-)}, partitioned into five splits. The key principle of the data set is to capture the images as close to real world conditions as possible, including all variations in appearance, pose, lighting condition and image quality, to name a few. These conditions provide for an unconstrained and challenging learning problem: The first results on the Adience benchmark achieved $45.1\%$ accuracy for age classification and $77.8\%$ accuracy for gender classification using a  pipeline including a robust, (un)certainty based in-plane facial alignment step, Local Binary Pattern (LBP) descriptors, Four Patch LBP descriptors and a dropout-SVM classifier \cite{eidinger2014age}. For reference, the same classification pipeline achieves $66.6\%$ accuracy for age classification and $88.6\%$ accuracy for gender classification on the Ghallagher data set.
The authors of \cite{hassner2015effective} introduce a 3D landmark-based alignment preprocessing step, which computes frontalized versions of the unconstrained face images from \cite{eidinger2014age}, which slightly increases gender classification accuracy to $79.3\%$ on the Adience data set, otherwise using the same classification pipeline from \cite{eidinger2014age}.

The first time a DNN model was applied to Adience benchmark was with \cite{levi2015age}. The authors did resort to an end-to-end training regime, \eg the face frontalization preprocessing from \cite{hassner2015effective} was omitted and the model was completely trained from scratch, in order to demonstrate the feature learning capabilities of the neural network type classifier. The architecture used in \cite{levi2015age} is very similar to the BVLC Caffe Reference Model \cite{jia2014caffe}, with the fourth and fifth convolution layers being removed. The best reported accuracy ratings increased to $50.7\%$ for age classification and $86.6\%$ for gender classification, using an over-sampling prediction scheme with 10 crops taken from a sample (4 from the corners and the center crop, plus mirrored versions) instead of only the sample by itself \cite{levi2015age}.

To the best of our knowledge, the current state of the art results for age and gender predictions are reported in \cite{rothe2016deep} and \cite{dehghan2017dager} with $64\%$ and $91\%$ accuracy respectively. The model from \cite{rothe2016deep} was the winner of the ChaLearn Looking at People 2015 challenge \cite{escalera2015chalearn} and uses the VGG-16 layer architecture \cite{simonyan2014very}, which has been pretrained on the IMDB-WIKI face data set. This data set was also introduced in \cite{rothe2016deep} and is comprised of 523,051 labelled face images collected from IMDb and Wikipedia. Prior to pretraining on the IMDB-WIKI data, the model was initialized with the weights learned for the ImageNet 2014 challenge \cite{russakovsky2015imagenet}. The authors attribute the success of their model to large amounts of (pre)training data, a simple yet robust face alignment preprocessing step (rotation only), and an appropriate choice of network architecture. 

The $91\%$ accuracy achieved by the commercial system from \cite{dehghan2017dager} is supposedly backed by 4,000,000 carefully labelled but non-public training images.
The authors identify their use of landmark-based facial alignment preprocessing as a critical factor to achieve the reported results.
Unfortunately no details are given about the model architecture in use. The authors of \cite{dehghan2017dager} compare their results to \cite{rothe2016deep} and other systems, yet only selectively list the age estimation of competing methods, such as \cite{rothe2016deep}.
The authors of \cite{dehghan2017dager} also report the gender recognition performance of \cite{rothe2016deep} as only $88.75\%$, which is rather low given the early results from \cite{levi2015age}, the performance of \cite{rothe2016deep} on age recognition and our own attempts to replicate the models of referenced studies.

\begin{table}
\begin{center}
\begin{tabular}{ll|r|rl}
&& gender & age & age (1-off) \\
\hline
\cite{eidinger2014age}		&(2014) 	& 77.8 	& 45.1 	& 79.5 \\
\cite{hassner2015effective}	&(2015)		& 79.3 	& -- 	& -- \\
\cite{levi2015age}			&(2015) 	& 86.8 	& 50.7 	& 84.7 \\
\cite{rothe2016deep}		&(2016) 	& -- 	& 64.0 	& 96.6 \\
\cite{dehghan2017dager} 	& (2017) 	& 91.0 	& 61.3 	& -- \\
\hline
\end{tabular}
\end{center}
	\caption{An overview over the developments for age and gender recognition results on the Adience benchmark in recent years. Accuracy values are reported in percent.
}
	\label{tab:overview}
\end{table}

Recapitulating, we can identify three major factors contributing to the performance improvements among the models listed in Table \ref{tab:overview}: (1)~Changes in architecture. (2)~Prior knowledge via pretraining. (3)~Optional dataset preparation via alignment preprocessing.

In the following sections, this paper will briefly describe a selection of DNN architectures and investigate the influence of random weight initialization against pretraining on generic  (ImageNet) or task-specific (IMDB-WIKI) real world data sets, as well as the impact of data preprocessing by comparing affine reference frame based alignment techniques to coarse rotation-based alignment. Due to its size and the unconstrained nature of the data and the availability of previous results, we use the Adience benchmark data set as an evaluation sandbox. The dataset is available as only rotation aligned version, and as a version with images preprocessed using the affine in-plane alignment \cite{eidinger2014age}, putting the shown faces closer to a reference frame of facial features. We then use Layer-wise Relevance Propagation (LRP) \cite{bach2015pixel} to give a glimpse into the model's prediction strategy, visualizing the facial features used for prediction on a per-sample basis in order to explain major performance differences.

\section{Architectures, Preprocessing and Model Initialization}
\label{sec:modelarchitectures}

This section provides an overview about the evaluated DNN architectures, data preprocessing techniques and weight initialization choices. All models are trained using the Caffe Deep Learning Framework \cite{jia2014caffe},
with code based on \url{https://github.com/GilLevi/AgeGenderDeepLearning},
containing the configurations to reproduce the results from  \cite{levi2015age}.

\subsection{Evaluated Models}
We compare the architectures of the model used in \cite{levi2015age} (in the following referred to as AdienceNet), the BVLC Caffe Reference Model \cite{jia2014caffe} (or short: CaffeNet), the GoogleNet \cite{szegedy2015going} and the VGG-16 \cite{simonyan2014very}, on which state of the art performance on age classification has been reported in \cite{rothe2016deep}. The AdienceNet is structurally similar to the CaffeNet, with the main difference lying in smaller convolution masks learned in the input layer ($7\times 7$ vs $11\times 11$) and two less convolution layers being present. The number of hidden units composing the fully connected layers preceding the output layer is considerably lower ($512$ vs $4096$) for AdienceNet. The VGG-16 consists of 13 convolution layers of very small kernel sizes of 2 and 3, which are interleaved with  similarly small pooling operations, followed by two fully connected layers with 4096 hidden units each, and a fully connected output layer.
The fourth model we use and evaluate is the GoogleNet, which connects a series of inception layers.
Each inception layer realizes multiple convolution/pooling sequences of different kernel sizes (sizes $3\times 3$ to $7\times 7$ in the input inception module) in parallel, feeding from the same input tensor, of which the outputs are then concatenated along the channel axis. Compared to the VGG-16 architecture, the GoogleNet is fast to train and evaluate, while slightly outperforming the VGG-16 model on the ImageNet 2014 Challenge with $6.6\%$ vs $7.3\%$ top-5 error in the classification task \cite{russakovsky2015imagenet}.

\subsection{Data Preprocessing}
One choice to be made for training and classification is regarding data preprocessing.
The SVM-based system from \cite{hassner2015effective} improves upon \cite{eidinger2014age} by introducing a 3D face frontalization preprocessing step, with the goal of rendering the inputs to the pipeline invariant to changes in pose.
Landmark-based preprocessing also is identified in \cite{dehghan2017dager} as an important step for obtaining the reported model performances.
Both \cite{levi2015age} and \cite{rothe2016deep} only employ simple rotation based preprocessing, which roughly aligns the input faces horizontally, trusting the learning capabilities of neural networks to profit from the increased variation in the data and learn suitable data representations.

The Adience benchmark data set provides both a version of the data set with images roughly rotated to horizontally aligned faces, as well as an affine 2D in-plane aligned version for download. We prepare training and test sets from both versions using and adapting the original splits and data preprocessing code for \cite{levi2015age} available for download on github. We also create a mixed data set from a union of both previous data sets, which has double the number of training samples and allows the models to be trained on both provided alignment techniques simultaneously.

\subsection{Weight Initialization}
An invaluable benefit of DNN architectures is the option to use pretrained models as a starting point for further training.
Compared to random weight initialization, using a pretrained models as starting points often results in faster convergence and overall better model results, due to initializing the model with meaningful filters.

In this paper, we compare models initialized with random weights to models starting with weights trained on other data sets, namely the ImageNet data set and the IMDB-WIKI data sets, whenever model weights are readily available. That is, we try to replicate the results from \cite{levi2015age} and train an AdienceModel only from scratch, since no weights for either pretraining data set are available. Instead, we use the comparable CaffeNet to estimate the results obtainable when initialzing the model with ImageNet weights. We also train the GoogleNet from scratch and initialized with ImageNet weights. Due to the excessive training time required for the VGG-16 model, we only try to replicate the results from \cite{rothe2016deep} and train models both initialized with available ImageNet and IMDB-WIKI weights.

\section{Visualizing Model Perception}
\label{sec:visualizingmodelperception}

We complement our quantitative analysis
in Section \ref{sec:evaluationandresults} with qualitative insights on the perception and reasoning of the models
by explaining the predictions made via the importance of features for or against a decision at input level. Following the success of DNNs, the desire to understand the inner workings of those black box models has vitalized research efforts dedicated to increasing the transparency of complex models. Several methods for explaining individual  predictions have emerged since then, with robust yet computationally expensive occlusion-based \cite{zeiler2014visualizing} and sampling-based analysis \cite{ribeiro2016should, zintgraf2016new}, (gradient-based) sensitivity analysis \cite{gevrey2003review,baehrens2010explain,simonyan2013deep} and backpropagation-type approaches \cite{bach2015pixel,zeiler2014visualizing,MonPR17} among them. 
In an intensive study \cite{samek2016evaluating}, Layer-wise Relevance Propagation (LRP) was found to outperform considered competing approaches in computing meaningful explanations for decisions made by DNN classifiers. Further, the method is in contrast to sampling or occlusion-based approaches computationally inexpensive and applicable to a wide range of architectures and classifier types \cite{bach2015pixel,lapuschkin2016analyzing}.
We therefore use LRP to supportively complement the quantitative results shown in Section \ref{sec:evaluationandresults} and visualize the perception of the model and its interaction with the input under the evaluated training conditions. For our experiments, we use the current version\footnote{\url{https://github.com/sebastian-lapuschkin/lrp_toolbox/tree/caffe-wip}} of the toolbox \cite{lapuschkin2016lrp} provided by the authors.

We refer the interested reader to \cite{MonArXiv17} for a tutorial on methods for understanding and interpreting deep neural networks.

\subsection{Layer-wise Relevance Propagation for DNNs}
\label{sec:lrp}

LRP is a principled and general approach to decompose the output of a decision function $f$, given an input $x$, into so-called \emph{relevance values} $R_p$ for each component $p$ of $x$ such that $\sum_p R_p = f(x)$.
The method operates iteratively from the model output to its inputs layer-by-layer in a backpropagation-style algorithm, computing relevance scores $R_i$ for hidden units in the interim. Each $R_i$ corresponds to the contribution an input or hidden variable $x_i$ has had to the final prediction, such that $f(x) = \sum_i R_i$ is true for all layers.
The method assumes that the decision function of a model can be decomposed as a feed-forward graph of neurons, \eg
\begin{equation}
x_j = \sigma\left(\sum_i x_iw_{ij} + b_j\right),
\end{equation}
where $\sigma$ is some monotonically increasing nonlinear function (\eg a ReLU), $x_i$ are the neuron inputs, $x_j$ is the neuron output and $w_{ij}$ and $b_j$ are the learned weight and bias parameters. The behaviour of LRP can be described by taking as example a single neuron $j$: That neuron receives a relevance quantity $R_j$ from neurons of the upper layer, which is to be redistributed to its input neurons $i$ in the lower layer, proportionally to the contribution of $i$ in the forward pass:
\begin{equation}
R_{i\leftarrow j} = \frac{z_{ij}}{z_j} R_j
\label{eq:lrpdecompose}
\end{equation}
Here, $z_{ij}$ is a quantity measuring the contribution of neuron $i$ to the activation of neuron $j$ and $z_j$ is the aggregation of all forward messages $z_{ij}$ over $i$ at $j$. 
The relevance score $R_i$ at neuron $i$ is then consequently obtained by pooling all incoming relevance quantities $R_{i\leftarrow j}$ from neurons $j$ to which $i$ contributes:
\begin{equation}
R_i = \sum_j R_{i \leftarrow j}
\label{eq:lrppool}
\end{equation}
Both the above relevance decomposition and pooling steps satisfy a local conservation property, \ie
\begin{equation}
R_i = \sum_j R_{i \leftarrow j}~~~\text{and}~~~\sum_i R_{i\leftarrow j} = R_j
\label{eq:lrpconservation}
\end{equation}
ensuring $f(x) = \sum_i R_i$ for $i$ iterating over the neurons of any layer of the network.

The relevance redistribution obtained from Equations \ref{eq:lrpdecompose} and \ref{eq:lrppool} is a very general one, with exact definitions depending on a neuron or input's type and position in the pipeline \cite{lapuschkin2016analyzing}. All DNN models considered in this paper consist in one part of ReLU-activated (convolutional) feature extraction layers towards the bottom, followed by inner product layers serving as classifiers \cite{NIPS2010_4061}. We therefore apply to inner product layers the $\epsilon$-decomposition
\begin{equation}
R_{i\leftarrow j} = \frac{x_iw_{ij}}{b_j + \sum_i x_iw_{ij}} R_j
\end{equation}
with small epsilon ($\epsilon=0.01$) of matching sign added to the denominator for numeric stability, to truthfully represent the decisions made via the layers' linear mappings consistently. 
Since the ReLU activations of the convolutional layers below serve as a gate to filter out weak activations, we apply the $\alpha\beta$ decomposition formula with $\beta=-1$ \cite{bach2015pixel}
\begin{equation}
R_{i\leftarrow j} = \left(\alpha\frac{z^+_{ij}}{\sum_iz^+_{ij}} + \beta\frac{z^-_{ij}}{\sum_iz^-_{ij}}\right)R_j,
\end{equation}
which handles the activating and inhibiting parts of $z_{ij}$ separately as $z^+_{ij}$ and $z^-_{ij}$ and weights them with $\alpha$ and $\beta$ respectively \cite{bach2015pixel}. Since $z_{ij} = z^+_{ij} + z^-_{ij}$, enforcing $\alpha+\beta=1$ ensures the conservation property from Equation \ref{eq:lrpconservation}. Theoretical insights into above decomposition types can be found in \cite{MonPR17}.

Once relevance scores are obtained on (sub)pixel level, we sum-pool the relevance values over the color channel axis. This leaves us with only one value $R_p$ per pixel $p$. We visualize the results using a color map centered at zero, since $R_p \approx 0$ indicates neutral or no contribution of input component $p$ to $f(x)$ and $R_p > 0$ and $R_p < 0$ identify components locally speaking for or against the global prediction. All models use vastly different filter sizes (from 2 to 11) in the bottom layers. We follow \cite{BacICIP16} in distributing $R_j$ for all neurons of some of the lower layers uniformly across their respective inputs, such that the granularity of the visualizations for all models are comparable.

\section{Evaluation and Results}
\label{sec:evaluationandresults}

We score all trained models using the oversampling evaluation scheme \cite{levi2015age}, by using the average prediction from ten crops (four corner and one center crop, plus mirrored versions) per sample. Results for age and gender prediction are shown in Tables \ref{tab:ageresultsoversampling} and \ref{tab:genderresultsoversampling} respectively. The columns of both tables correspond to the described models; the \textbf{A}dienceNet, \textbf{C}affeNet, \textbf{G}ooglenet and \textbf{V}GG-16. Following previous work we also report 1-off accuracy results -- the accuracy obtained when predicting at least the age label adjacent to the correct one -- for the age prediction task.

The row headers describe the training and evaluation setting: A first value of $[$i$]$ signifies the use of $[$i$]$n-plane face alignment from \cite{eidinger2014age} as a preprocessing step for training and testing, $[$r$]$ stands for $[$r$]$otation based alignment and $[$m$]$ describes results obtained when both rotation aligned and in-plane aligned images have been $[$m$]$ixed for training and images from the $[$r$]$ test set have been used for evaluation. Second values $[$n$]$ or $[$w$]$ describe weight initialization using Image$[$n$]$et  and IMDB-$[$w$]$IKI respectively. No second value means the model has been trained from scratch with random weight initialization.

\begin{table}
\begin{center}
\begin{tabular}{l|rrrr}
	& \textbf{A} & \textbf{C} & \textbf{G}& \textbf{V}\\
\hline
	 $[$i$]$ 		& 51.4 \tiny{87.0}	& 52.1 \tiny{87.9}	& 54.3 \tiny{89.1}	& -- \\
	 $[$r$]$ 		& 51.9	\tiny{87.4} & 52.3	\tiny{88.9} & 53.3	\tiny{89.9} & -- \\
	 $[$m$]$ 		& 53.6 \tiny{88.4}	& 54.3	\tiny{89.7} & 56.2	\tiny{90.7} & -- \\
	 \hline
	 $[$i,n$]$ 	& --  	& 51.6 \tiny{87.4}	& 56.2 \tiny{90.9}	& 53.6 \tiny{88.2} \\
	 $[$r,n$]$ 	& -- 	& 52.1 \tiny{87.0} 	& 57.4 \tiny{91.9}	& -- \\
	 $[$m,n$]$ 	& --	& 52.8 \tiny{88.3}	& 58.5 \tiny{92.6}	& 56.5 \tiny{90.0} \\
	 \hline
	 $[$i,w$]$ 	& -- 	& -- 	& -- 	& 59.7 \tiny{94.2} \\
	 $[$r,w$]$ 	& -- 	& --	& --	& --  \\
	 $[$m,w$]$ 	& -- 	& -- 	& -- 	& 62.8 \tiny{95.8} \\
\hline
\end{tabular}
\end{center}
	\caption{Result for \textbf{age} classification in accuracy in percent, using oversampling for prediction. Small numbers next to the accuracy score show 1-off, e.g the accuracy with which at least an adjacent age group has been predicted.}
	\label{tab:ageresultsoversampling}
\end{table}

\begin{table}
\begin{center}
\begin{tabular}{l|rrrr}
	& \textbf{A} & \textbf{C} & \textbf{G}		& \textbf{V}\\
\hline
	 $[$i$]$ 		& 88.1 	& 87.4	& 87.9		& -- \\
	 $[$r$]$ 		& 88.3 	& 87.8 	& 88.9		& -- \\
	 $[$m$]$ 		& 89.0 	& 88.8 	& 89.7 		& -- \\
	 \hline
	 $[$i,n$]$ 	& -- 	& 89.9 	& \textbf{91.0} & \textbf{92.0} \\
	 $[$r,n$]$ 	& --	& 90.6 	& \textbf{91.6}	& -- \\
	 $[$m,n$]$ 	& --	& 90.6 	& \textbf{91.7}	& \textbf{92.6}\\
	 \hline
	 $[$i,w$]$ 	& --	& --	& --			& 90.5\\
	 $[$r,w$]$ 	& --	& --	& --			& -- \\
	 $[$m,w$]$ 	& --	& --	& --			& \textbf{92.2}\\
\hline
\end{tabular}
\end{center}
	\caption{Results for \textbf{gender} classification in accuracy, using oversampling for prediction. Bold values match or exceed the currently reported state of the art results from \cite{dehghan2017dager} on the Adience benchmark.}
	\label{tab:genderresultsoversampling}
\end{table}

\begin{figure*}
\begin{center}
\includegraphics[width=0.9\textwidth]{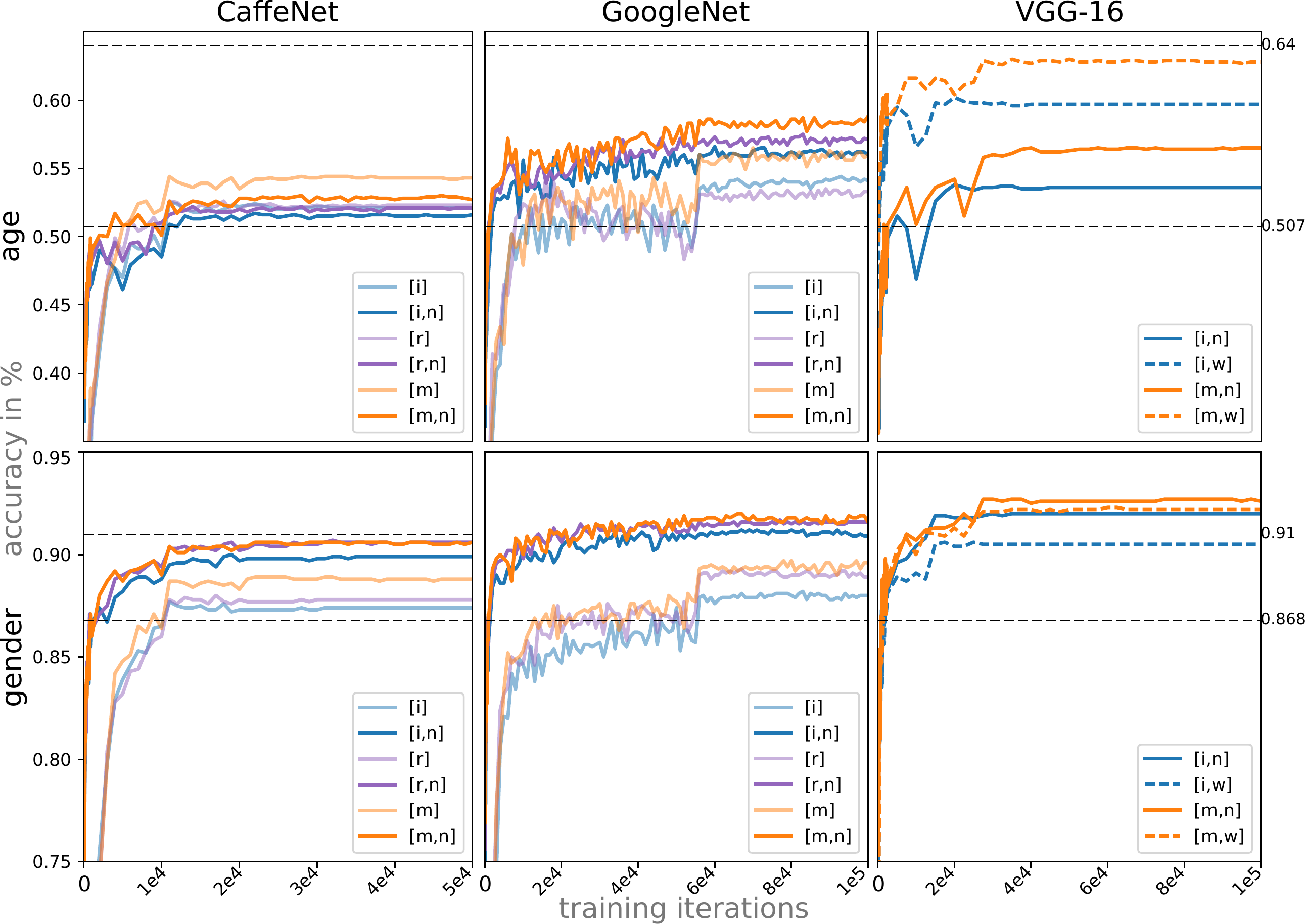}
\end{center}
	\caption{The plots are ordered column-wise over model architectures and row-wise according to prediction problem, showing model performance over training time given different initializations and data preprocessing settings.
	The top and bottom dashed lines in each plot show worst and best reference accuracy results from \cite{levi2015age,rothe2016deep} and \cite{dehghan2017dager}, with the horizontal axis increasing with training iterations.
	Thick lines show results taken by us. Color coding corresponds to data preprocessing and shading to model initialization: Blue color stands for affine $[$i$]$n-plane alignment. Violet lines correspond to $[$r$]$otation alignment. Orange lines show the model performances for training on the $[$m$]$ixed training set. Translucent line color stands for training with random model initialization, fully opaque and solid lines show performance for finetuning on ImageNet weights and dashed lines correspond to model initialization using IMDB-WIKI weights, only applied to the VGG-16 model.
	All results are averaged over the five splits of the Adience data set.} 
	\label{fig:trainingprogress}
\end{figure*}

The results in above tables list the measured performance after a fixed amount of training steps. Intermediate models which might have shown slightly better performance are ignored in favour of comparability.
With our attempt to replicate the results from \cite{levi2015age} based on the code provided by the authors, we managed to exceed the reported results in both accuracy by ($+1.2\%$) and 1-off accuracy ($+2.7\%$) for age prediction and accuracy ($+1.5\%$) for gender prediction. As expected, the structurally comparable CaffeNet architecture obtains relatable results for both learning problems with random model weight initialization. We then further compared the relatively fast to train CaffeNet model to the GoogleNet model in all data  preprocessing configurations when trained from scratch and fine-tuned based on the ImageNet weights. We try to replicate the measurements from \cite{rothe2016deep} to verify the observations made based on the other models. Here, we did not fully manage to reach the reported results, despite using the model pre-trained on the IMDB-WIKI data as provided by the authors. However, we closely scrape by the reported results with slight differences in both accuracy ($-1.2\%$) and 1-off accuracy ($-0.8\%$), averaged over all five splits of the data with a model trained on the mixed training set.
In all evaluated settings shown in Figure \ref{fig:trainingprogress} we can observe overall trends in choices for architecture, dataset composition and preprocessing and model initialization. 

\subsection{Remarks on Model Architecture}
\begin{figure}
\begin{center}
\includegraphics[width=0.45\textwidth]{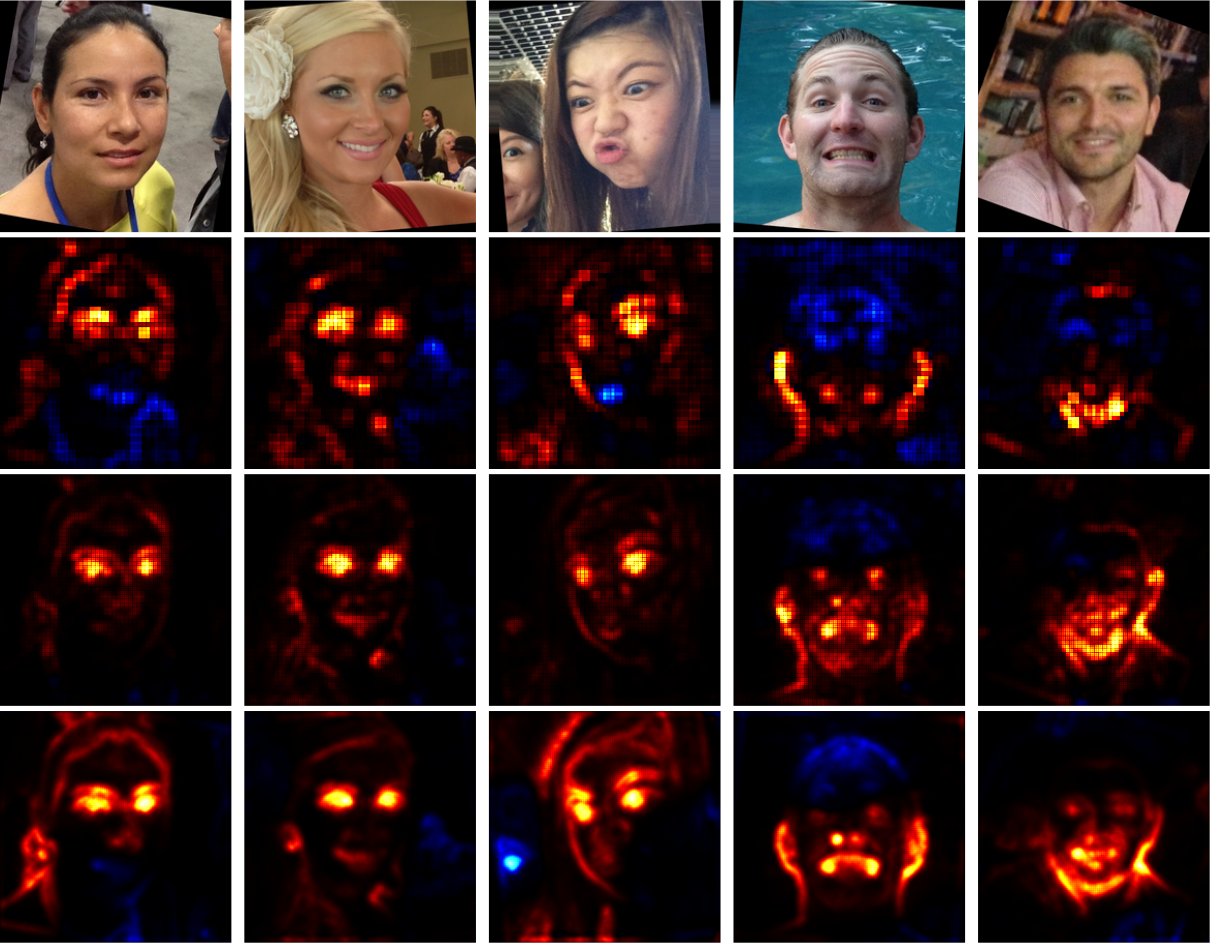}
\end{center}
\caption{From top to bottom: Input image, followed by relevance maps for the best performing \textbf{CaffeNet}, \textbf{GoogleNet} and the \textbf{VGG-16} model for \textbf{gender} prediction.
Hot colors identify parts of the image contributing to the predicted class. Cold hues show evidence contradicting the predicted class, as perceived by the model. Smoother heatmaps are a consequence of smaller filters and stride in the bottom layers.}
\label{fig:howarchitecturesusefaces}
\end{figure}

In all settings, the CaffeNet architecture is outperformed by the more complex and deep GoogleNet and VGG-16 models. For gender classification under comparable settings, the best VGG-16 models outperform the best GoogleNet models. Figure \ref{fig:howarchitecturesusefaces} visualizes the different characteristics of input faces as used by the classifiers to predict either male or female gender.

We observe that model performance correlates with network depth, which in turn correlates with the structure observable in the heatmaps computed with LRP. For instance, all models recognize female faces dominantly via hair line and eyes, and males based on the bottom half of the face. The CaffeNet model tends to contentrate more on isolated aspects of a given input compared to the other two, especially for men, while being less certain in its prediction, reflected by the stronger negative relevance.

\subsection{Observations on Preprocessing}
\begin{figure*}
\begin{center}
\includegraphics[width=.95\textwidth]{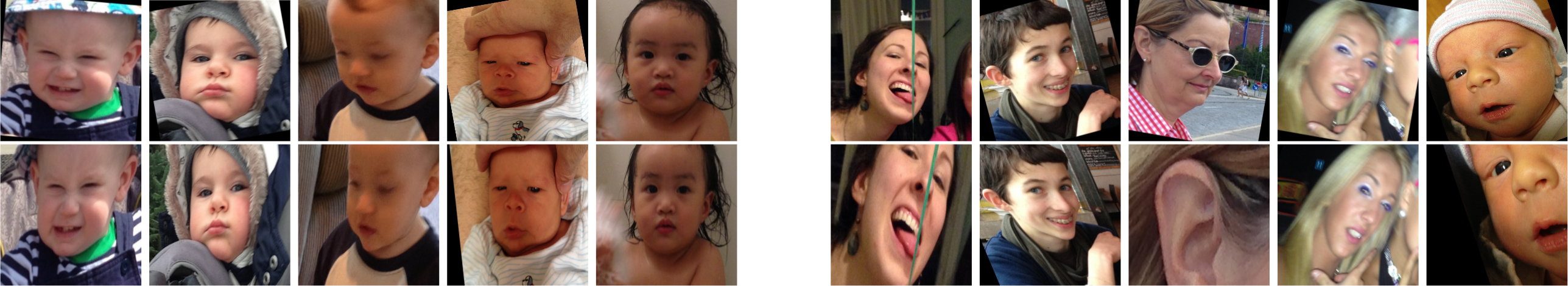}
\end{center}
	\caption{
Top: Samples taken from the only rotation aligned variant of the Adience data set. Bottom: In-plane aligned samples. The left five image pairs show faces taken from the age group of (0-2) which are classified correctly under rotation alignment and are placed at least one age group above by the predictor under landmark-based alignment, with the middle image to the left being predicted as age group (8-13) by the GoogleNet. The in-plane alignment technique applied to one variant of the Adience data set tends to elongate faces vertically. The remaining image show misclassified and misaligned samples picked at random.}
	\label{fig:facesrotatedvsaligned}
\end{figure*}

For all three models, we can observe the overall trend for both prediction problems, that the in-plane alignment preprocessing step is not beneficial to classifier performance, compared to rotation alignment.
The only exception to this trend is the randomly initialized GoogleNet model, which loses one percent accuracy for age prediction under rotation alignment albeit still gaining performance in measured 1-off prediction. We reason the better performance on only rotation aligned images to be justified in the potential of and for DNNs to learn for the domain of face images canonically meaningful sets of features. For the face images aligned using the technique presented in \cite{eidinger2014age}, this is more difficult. Especially for images of children, the faces aligned to reference frames suitable for adults result in head shapes of uncharacteristic aspect ratios for the age group or even faulty alignments.
Figure \ref{fig:facesrotatedvsaligned} demonstrates the nature of this artefactual noise introduced to the data by unsuitable alignment.

All models benefit the most from combining both the rotation aligned and the landmark aligned data sets for training.
For one, this effectively doubles the training set sizes,
but also -- perhaps more importantly -- allows the learning of a more robust feature set:
The models trained on a combination of both the landmark aligned and rotation aligned images perform well on test sets resulting from both preprocessing techniques.
Tables \ref{tab:ageresultsoversampling} and \ref{tab:genderresultsoversampling} show results for models trained on the combined set which were evaluated on the rotation aligned test set.

Performance measurements on in-plane aligned data are with $<1\%$ only insignificantly lower.

In order to underline the effect of increased robustness of the models trained on the more diverse $[$r$]$oration aligned training set we evaluated models trained on $[$i$]$n-plane aligned images with $[$r$]$otation aligned test images and vice versa. Corresponding model performances are listed in Tables \ref{tab:ageresultsopposite} and \ref{tab:genderresultsopposite}.
Some models trained on data prepared with one alignment technique evaluated against the test set of the other perform even worse than the early SVM-based models from \cite{eidinger2014age}, despite their competitive results from the combined training set.  
The models trained on the
in-plane aligned images have more difficulty predicting on the unseen setting than the models trained on the only rotated images, where the original facial pose and the proportions of the face image are mostly preserved.

For the VGG-16 model, we compared the in-plane alignment to the mixed training set -- the worst to the best expected results. Here again, the mixed training data results in a better model than when only in-plane alignment is used.
Figure~\ref{fig:trainingprogress} shows an overview of all results over training time.

\subsection{Observations on Initialization}

\begin{figure}
\begin{center}
\includegraphics[width=0.45\textwidth]{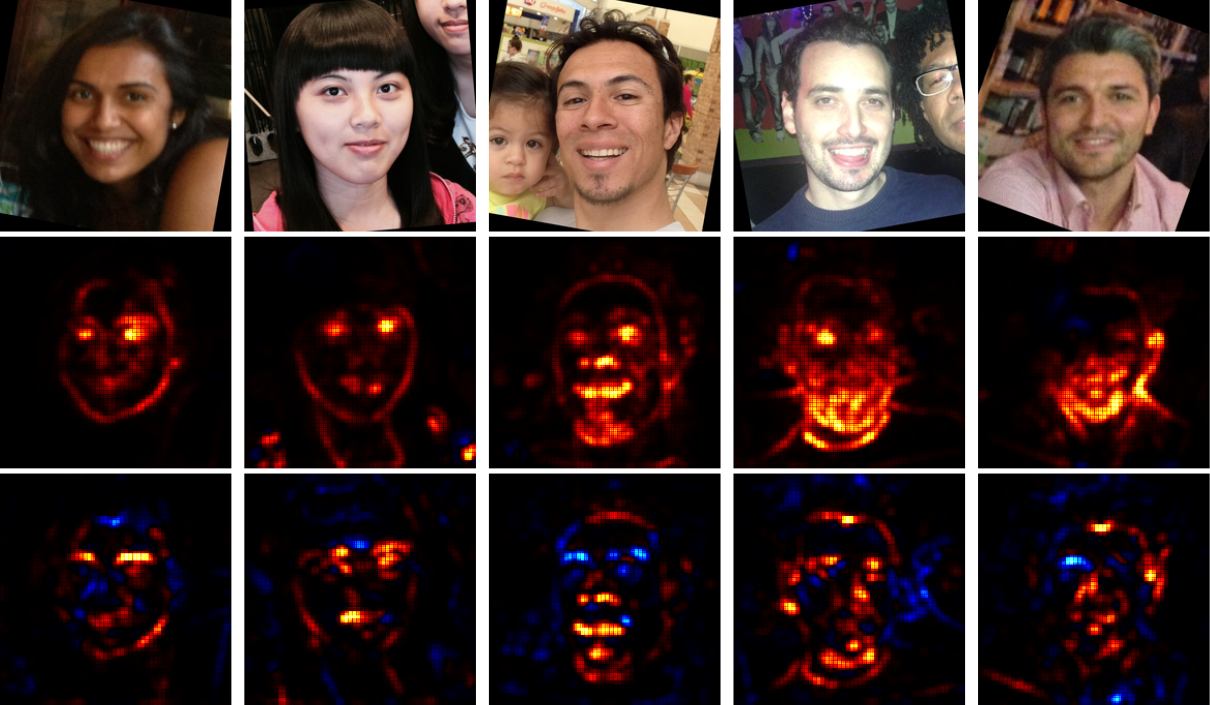}
\end{center}
\caption{Heatmaps for \textbf{GoogleNet} models and \textbf{gender} recognition. Input images are shown above heatmaps for a DNN pretrained on Imagenet, which are shown above heatmaps for a DNN initialized randomly. The finetuned model predicts based on an ensemble of facial features, whereas the model starting with random weights has overfit on an isolated set of features characteristic to the target classes.}
\label{fig:googlenetinitgender}
\end{figure}

\begin{figure}
\begin{center}
\includegraphics[width=0.45\textwidth]{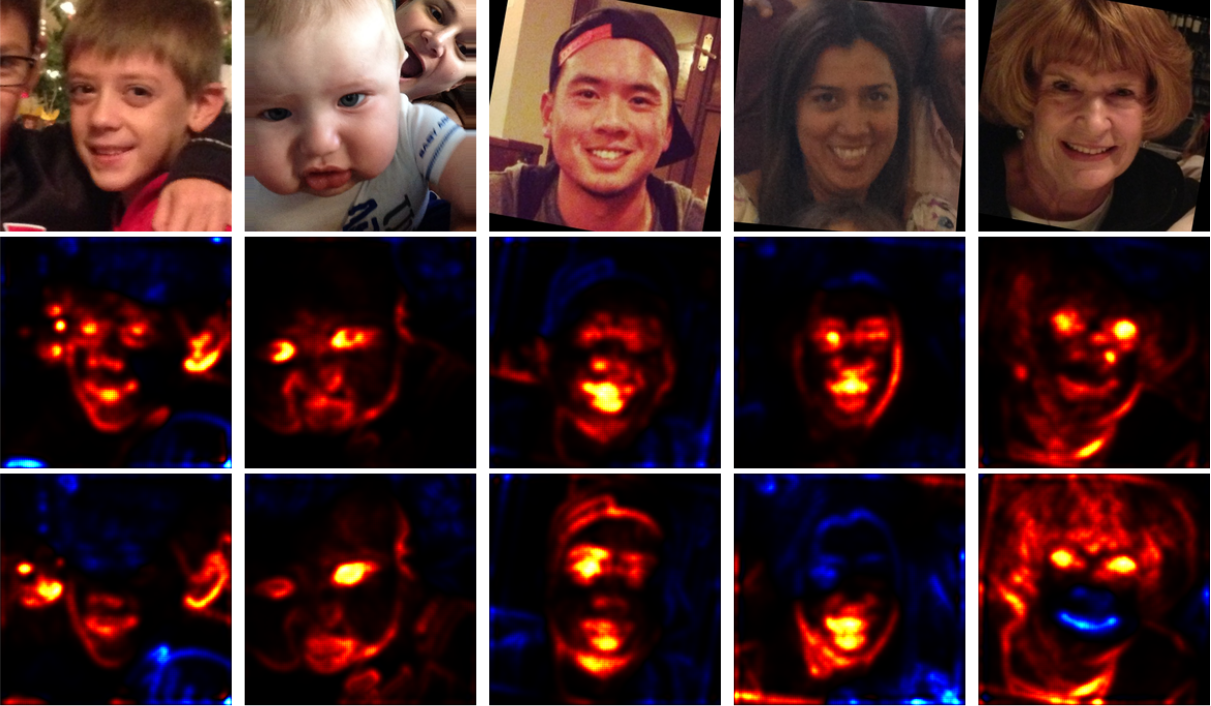}
\end{center}
\caption{Heatmaps for \textbf{VGG-16} and \textbf{age} prediction. Input images are shown above heatmaps for a DNN pretrained on IMDB-WIKI, which are shown above heatmaps for a DNN pretrained on ImageNet.}
\label{fig:vgginitage}
\end{figure}

We find that the GoogleNet model responds well to fine-tuning on the weights pre-trained on ImageNet and responds with an increase in performance for both classification problems and in all dataset configurations. The CaffeNet, however, slightly loses performance when fine tuned for age group prediction, while benefiting in gender prediction. The better response of the GoogleNet compared to the CaffeNet, when initialized with their respective ImageNet weights might be caused by the quality of the initial parameters: While the GoogleNet achieves a $6.6\%$ top-5 error on ImageNet, the CaffeNet only reaches $19.6\%$.
Evaluating on the \textit{incorrect} test data (Tables \ref{tab:ageresultsopposite} and \ref{tab:genderresultsopposite}), both fine tuned models trained on rotation aligned images manage to recover their respective performance ratings compared to models trained from scratch and being evaluated on the \textit{correct} data. The GoogleNet model even exceeds the performance of the same architecture initialized randomly but both trained and evaluated on the rotated images.
The measurable beneficial effect of appropriate pretraining  is visualized in Figures~\ref{fig:googlenetinitgender} and \ref{fig:vgginitage}.
ImageNet pretraining leads to the use of larger and meaningful parts of the face for prediction for the GoogleNet, while the randomly initialized model picks out single characteristics during training which correlate the most with the target class.
This includes eyebrows and lips defining female faces and nose, chin and uncovered ears for men for gender recognition.
We see comparable results for the VGG-16 on age group estimation when comparing pretraining on ImageNet and IMDB-WIKI.
The model initialized with IMDB-WIKI weights, with the pretraining task being age estimation on 101 age categories, concentrates more on the facial features themselves, while the ImageNet-initialized one is more prone to distraction from background elements and clothing items.
Facial features seen in examples of opposing classes of the respectively weaker models in both figures -- independent of the ensemble of facial features -- leads to less certain, noisy decisions.
For the problem of gender recognition, the VGG-16 is affected less from weight initialization than from the quality of data preprocessing. Here, IMDB-WIKI pretraining might have an only diminished effect due to firstly the ImageNet weights providing an already good set of starting weights and secondly, the pretraining objective (age recognition) being orthogonal to the task of gender recognition. In fact, other than for age recognition, the VGG-16 models initialized with ImageNet weights converged to better parameters than their counterparts.

Figure \ref{fig:trainingprogress} reports the prediction performances of the CaffeNet, the GoogleNet and the VGG-16 model in all evaluated settings, averaged over the five splits of the Adience data set. The recorded model scores over time illustrate that suitably initializing a model largely outweighs the problems introduced with artefactual data in our experiments. Next to the overall better model obtained after convergence, we also observe a considerably faster increase in the learning progress early in training.

\begin{table}
\begin{center}
\begin{tabular}{l|rrr}
	& \textbf{A} & \textbf{C} & \textbf{G}\\
\hline
	 $[$i$]$ 		& 40.8 \tiny{75.4}	& 40.3 \tiny{76.3}	& 44.6 \tiny{80.8}	\\
	 $[$r$]$ 		& 46.9	\tiny{82.8} & 46.1	\tiny{82.5} & 46.4	\tiny{83.2}  \\
	 \hline
	 $[$i,n$]$ 	& --  	& 45.2 \tiny{82.02}	& 49.4 \tiny{87.2}	\\
	 $[$r,n$]$ 	& -- 	& 48.8 \tiny{84.9} 	& 53.6 \tiny{89.9}	\\
\hline
\end{tabular}
\end{center}
	\caption{
	Test set swapping results for \textbf{age} prediction.
	Performance is considerably worse when the \textit{incorrect} preprocessing is used for testing, due to overfit feature sets. Pretraining can yield robust model parameters, compensating for the deviating test statistics.
	}
	\label{tab:ageresultsopposite}
\end{table}

\begin{table}
\begin{center}
\begin{tabular}{l|rrr}
	& \textbf{A} & \textbf{C} & \textbf{G}\\
\hline
	 $[$i$]$ 		& 81.1	& 80.5	& 83.5\\
	 $[$r$]$ 		& 81.3 	& 84.6 	& 86.0\\
	 \hline
	 $[$i,n$]$ 	& -- 	& 84.5 	& 89.6 \\
	 $[$r,n$]$ 	& --	& 88.5 	& 90.0 \\
\hline
\end{tabular}
\end{center}
	\caption{
	Test set swapping results for \textbf{gender} prediction.
	}
	\label{tab:genderresultsopposite}
\end{table}

\section{Conclusion}
Recent deep neural network models are able to accurately analyze human face images, in particular recognize the persons' age, gender and emotional state.
Due to their complex non-linear structure, however, these models often operate as black-boxes and until very recently it was unclear {\it why} they arrived at their predictions.
In this paper we opened the black-box classifier using Layer-wise Relevance Propagation and investigated which facial features are actually used for age and gender prediction.
We compared different image preprocessing, model initialization and architecture choices on the challenging Adience dataset and discussed how they affect performance.
By using LRP to visualize the models' interactions with the given input samples, we demonstrate that appropriate model initialization via pretraining counteracts overfitting, leading to a holistic perception of the input.
With a combination of simple preprocessing steps, we achieve state of the art performance for gender classification on the Adience benchmark data set.

{\small
\bibliographystyle{ieee}
\bibliography{bibliography}
}

\end{document}